\documentclass[11pt]{article}

\usepackage[final]{acl}

\usepackage{times}
\usepackage{latexsym}
\usepackage{amssymb}
\usepackage{amsmath}
\usepackage{booktabs}
\usepackage{natbib}
\usepackage{array}
\usepackage{enumitem}
\usepackage{tabularx}
\usepackage{makecell}
\usepackage{multirow}
\usepackage[most]{tcolorbox}
\usepackage[table]{xcolor}
\usepackage{listings}

\lstdefinestyle{promptstyle}{
  basicstyle=\ttfamily\small,
  breaklines=true,
  breakindent=0pt,
  columns=fullflexible,
  keepspaces=true,
  frame=single,
  showstringspaces=false,
  xleftmargin=0.5em,
  framexleftmargin=0.5em
}

\lstdefinestyle{asciistyle}{
  basicstyle=\ttfamily\scriptsize,
  breaklines=false,
  breakindent=0pt,
  columns=fullflexible,
  keepspaces=true,
  frame=single,
  showstringspaces=false,
  xleftmargin=0.5em,
  framexleftmargin=0.5em
}

\usepackage[T1]{fontenc}

\usepackage[utf8]{inputenc}

\usepackage{microtype}

\usepackage{inconsolata}

\usepackage{graphicx}

\title{MCJudgeBench: A Benchmark for Constraint-Level Judge Evaluation in Multi-Constraint Instruction Following}

\author{
Jaeyun Lee\textsuperscript{1}\thanks{Corresponding author: \texttt{jaeyun.lee@cs.ox.ac.uk}} \quad
Junyoung Koh\textsuperscript{2} \quad
Zeynel Tok\textsuperscript{1} \quad
Hunar Batra\textsuperscript{1} \quad
Ronald Clark\textsuperscript{1} \\
\\
\textsuperscript{1}University of Oxford \quad
\textsuperscript{2}Yonsei University
}

\begin{document}
\maketitle

\begin{abstract}
Multi-constraint instruction following requires verifying whether a response satisfies multiple individual requirements, yet LLM judges are often assessed only through overall-response judgments. We introduce \textbf{MCJudgeBench}, a benchmark for constraint-level judge evaluation in multi-constraint instruction following. Each instance includes an instruction, a candidate response, an explicit constraint list, per-constraint gold labels in \{\textit{yes}, \textit{partial}, \textit{no}\}, and controlled response-side perturbations. The evaluation protocol further includes evaluation prompt variants to test judge stability. We evaluate proprietary and open-source LLM judges using both correctness and inconsistency metrics, distinguishing intrinsic inconsistency under stochastic decoding from procedural inconsistency under prompt and response perturbations. Our results show that judge reliability has multiple dimensions: strong overall performance does not guarantee equally reliable detection across label categories, especially for rarer \textit{partial} and \textit{no} cases. Judges with higher correctness do not always have lower inconsistency. Evaluation with reasoning improves correctness but does not uniformly improve stability. These findings motivate evaluating LLM judges at the constraint level to study these failure modes.
\end{abstract}

\section{Introduction}
Instruction following often involves satisfying several constraints at once, such as requirements on content, format, style, or length. Recent benchmarks show that performance declines as the number of constraints increases and they become harder to satisfy jointly, making prompt-level evaluation insufficient for understanding which requirements a model actually satisfies \citep{jiang2024followbench, wen2024complexbench}. In practice, however, assessing multi-constraint adherence at scale is expensive, which has encouraged the use of \emph{LLM-as-a-judge} for open-ended responses.

In parallel, there is increasing evidence that LLM judges are not always reliable. Their judgments can vary with prompt formulation and evaluation protocol, they may favor outputs that appear stylistically stronger despite weaker instruction adherence, and they can exhibit systematic biases under controlled changes \citep{zeng2024llmbar, liu2025reife, ye2024calm, li2025llmsreliablyjudgeyet, yang2026fairjudgeadaptivedebiasedconsistent}. Existing judge benchmarks have made progress on pairwise preferences, scalar judgments, and rubric-based evaluation \citep{tan2025judgebench, muller2025grouse}, but they do not directly address whether a judge can verify individual constraints in multi-constraint instruction following tasks.

This is particularly important in multi-constraint settings, where a response may satisfy some requirements while failing others. In such settings, evaluation must identify which constraints are satisfied or violated, rather than relying only on an overall score. A reliable evaluator must also remain stable under invariance perturbations to prompts and candidate response form. However, constraint-level judge stability in multi-constraint instruction following remains underexplored.

In this work, we introduce \textbf{MCJudgeBench}, a meta-evaluation benchmark for LLM-as-a-judge on multi-constraint instruction following. Building on ComplexBench and InFoBench \citep{wen2024complexbench, qin2024infobench}, the benchmark represents each instance as an instruction, a candidate response, and an explicit list of constraints, together with per-constraint adherence labels drawn from three classes: \textit{yes}, \textit{partial}, or \textit{no}. We further construct controlled invariance sets through candidate response variants that preserve the underlying adherence labels. In the evaluation protocol, we additionally introduce evaluation prompt variants to test judge stability under alternative formulations of the same verification task.

Beyond benchmark construction, we use this setting to study how current LLM judges behave on multi-constraint adherence verification. We evaluate judge correctness on the original instances and decompose judge instability into two sources: intrinsic inconsistency under stochastic decoding and procedural inconsistency under two invariance perturbation axes. Our results show that aggregate evaluation can hide substantial variation in constraint-level behavior, motivating systematic evaluation of judge reliability in this setting. Our contributions are as follows:

\begin{itemize}[leftmargin=1.2em, itemsep=2pt, topsep=1pt, parsep=0pt]

    \item MCJudgeBench, a new meta-evaluation benchmark for LLM-as-a-judge on multi-constraint instruction following, with per-constraint labels and controlled invariance sets.        
    \item An evaluation framework for judge reliability, covering correctness on original instances and two sources of judge instability: intrinsic inconsistency under stochastic decoding and procedural inconsistency under invariance perturbations.
    \item A baseline evaluation on the benchmark across proprietary and open-source LLM judges, showing how reliability varies across constraint types, constraint count, and perturbation conditions.
\end{itemize}

\section{Related Work}

\begin{table*}[t]
\centering
\small
\setlength{\tabcolsep}{5.5pt}
\renewcommand{\arraystretch}{1.12}
\begin{tabular}{lcccccc}
\toprule
\textbf{Benchmark} &
\textbf{Target} &
\textbf{IF} &
\makecell[c]{\textbf{Multi-}\\\textbf{const.}} &
\makecell[c]{\textbf{Decomposed} \\\textbf{eval.}} &
\makecell[c]{\textbf{Perturb.}\\\textbf{tests}} &
\makecell[c]{\textbf{Constraint-level}\\\textbf{judge eval.}} \\
\midrule
IFEval \citep{zhou2023ifeval} & Models & \checkmark & \checkmark & \checkmark & -- & -- \\ 
FollowBench \citep{jiang2024followbench} & Models     & \checkmark & \checkmark & \checkmark & --         & -- \\
ComplexBench \citep{wen2024complexbench} & Models     & \checkmark & \checkmark & \checkmark & --         & -- \\
InFoBench \citep{qin2024infobench}       & Models     & \checkmark & \checkmark & \checkmark & --         & -- \\
WildIFEval \citep{lior2025wildifeval}    & Models     & \checkmark & \checkmark & \checkmark & --         & -- \\
CFBench \citep{zhang2025cfbench}         & Models     & \checkmark & \checkmark & \checkmark & --         & -- \\
\midrule
LLMBar \citep{zeng2024llmbar}            & Evaluators & \checkmark & --         & --         & \checkmark & -- \\
ReIFE \citep{liu2025reife}               & Evaluators & \checkmark & --         & --         & --         & -- \\
GroUSE \citep{muller2025grouse}          & Evaluators & --         & --         & \checkmark & \checkmark & -- \\
\midrule
\textbf{MCJudgeBench}                       & Evaluators & \checkmark & \checkmark & \checkmark & \checkmark & \checkmark \\
\bottomrule
\end{tabular}

\caption{Comparison of \textbf{MCJudgeBench} with prior multi-constraint instruction following benchmarks and meta-evaluation benchmarks for automated evaluators and LLM judges. ``Decomposed eval.'' indicates evaluation beyond a single overall judgment, such as checking individual constraints, decomposed criteria, or rubric dimensions. ``Constraint-level judge eval.'' indicates meta-evaluation of whether a judge can verify individual constraints in multi-constraint instruction following.}

\label{tab:comparison}
\end{table*}

\paragraph{Instruction following and multi-constraint evaluation.} Recent work has moved instruction following evaluation beyond coarse prompt-level judgments toward a more fine-grained assessment of whether individual requirements are satisfied. IFEval \citep{zhou2023ifeval} focuses on objectively verifiable constraints, with each instruction containing one or more such constraints. FollowBench \citep{jiang2024followbench} extends this direction by constructing a controlled progression in difficulty over the same base instruction, incrementally adding constraints so that higher levels require satisfying a larger and more diverse set of requirements. Subsequent benchmarks broaden this line in different ways. ComplexBench \citep{wen2024complexbench} emphasizes the composition of multiple constraints and introduces a hierarchical taxonomy over constraint and composition types. InFoBench \citep{qin2024infobench} decomposes complex instructions into simpler criteria for more interpretable requirement-level evaluation. CFBench \citep{zhang2025cfbench} and WildIFEval \citep{lior2025wildifeval} extend evaluation to broader and more realistic constraint settings through systematic taxonomies, real-world scenarios, and real user instructions.

\paragraph{LLM-as-a-judge and meta-evaluation of evaluators.}
LLM-as-a-judge has become a common approach for scalable evaluation, but recent work shows that judge behavior is often sensitive to factors beyond the underlying quality of the response. LLMBar \citep{zeng2024llmbar} shows that judges can be misled by stylistic differences even when instruction adherence is weaker, while ReIFE \citep{liu2025reife} demonstrates that evaluation accuracy depends strongly on both the base LLM and the evaluation protocol. JudgeBench \citep{tan2025judgebench} and GroUSE \citep{muller2025grouse} further argue that judge quality cannot be assessed only through aggregate agreement or correlation, since evaluators may still fail on difficult objective comparisons or specific failure modes. Related work on bias analysis also finds persistent systematic preferences under controlled modifications \citep{ye2024calm}. Recent work further shows that LLM evaluators can be inconsistent and sensitive to prompt differences that are not meaningful to human evaluators \citep{stureborg2024large}. Evaluator consistency has also been studied through self-consistency and inter-scale consistency, showing that strong evaluator models are not necessarily consistent and that consistency should be considered when assessing LLM evaluators \citep{lee2025evaluating}.

\paragraph{Constraint-level judge evaluation.}
These two lines of work address complementary parts of the problem. Multi-constraint instruction following benchmarks study whether models satisfy complex sets of requirements, while meta-evaluation studies examine the reliability of evaluators through pairwise preferences, scalar judgments, or rubric-based scoring \citep{zheng2023judgingllmasajudgemtbenchchatbot, kim2024prometheus}. Prior work also shows the value of controlled perturbations for diagnosing evaluator behavior \citep{sai2021evaleval, ribeiro-etal-2020-beyond, hua2025flawartifactrethinkingprompt}, but this perspective has not been developed for constraint-level verification in multi-constraint instruction following. What remains underexplored is whether LLM judges can accurately and stably verify individual constraints, especially under invariance perturbations to the evaluation prompt or response form. Our work addresses this gap through MCJudgeBench, a benchmark for per-constraint judge evaluation under controlled invariance perturbations. Table~\ref{tab:comparison} summarizes the position of our benchmark relative to prior model-side and evaluator-side benchmarks.

\section{MCJudgeBench}
\label{sec:benchmark}

\subsection{Overview}
MCJudgeBench is a meta-evaluation benchmark for constraint-level judge evaluation in multi-constraint instruction following. Each instance is represented as
\[
x = (I, y, C, L, \mathcal{P}_y),
\]
where \(I\) denotes the instruction, \(y\) the natural candidate response, \(C = \{c_1,\dots,c_m\}\) the explicit constraint list, \(L = \{\ell_1,\dots,\ell_m\}\) the corresponding per-constraint gold labels with \(\ell_j \in \{\textit{yes}, \textit{partial}, \textit{no}\}\), and \(\mathcal{P}_y\) denotes an invariance perturbation set of candidate response variants. A label of \textit{yes} indicates that a constraint is clearly satisfied, \textit{no} indicates that it is clearly violated or missing, and \textit{partial} indicates partial but incomplete satisfaction. Figure~\ref{fig:mcjudge_example} shows a representative instance of MCJudgeBench.

\begin{figure}[t]
\centering
\footnotesize

\definecolor{yesgreen}{RGB}{0,120,60}
\definecolor{nored}{RGB}{170,40,40}
\definecolor{changeclr}{RGB}{70,90,140}

\begin{tcolorbox}[
    colback=gray!4,
    colframe=black!55,
    boxrule=0.5pt,
    arc=1mm,
    left=1mm,right=1mm,top=1mm,bottom=1mm,
    width=\columnwidth
]
\textbf{Instruction (\(I\))}\\
Write two sentences without using the letter \texttt{'e'}, ensuring grammatically correct sentences and valid word usage.

\vspace{0.6em}
\textbf{Candidate response (\(y\))}\\
\emph{Sunlight glows on a still pond. A fox hops near a lone tree.}

\vspace{0.6em}
\renewcommand{\arraystretch}{1.08}
\setlength{\tabcolsep}{4pt}
\begin{tabularx}{\linewidth}{@{}>{\raggedright\arraybackslash}X >{\raggedleft\arraybackslash}p{0.24\linewidth}@{}}
\textbf{Constraint \(c_j \in C\)} & \textbf{Label \(\ell_j \in L\)} \\
\midrule
1. Does the generated text consist of precisely two sentences?
& \textit{\textcolor{yesgreen}{yes}} \\

2. Does the generated text avoid using the letter \texttt{'e'}?
& \textit{\textcolor{nored}{no}} \\

3. Is the generated text grammatically correct with valid word usage?
& \textit{\textcolor{yesgreen}{yes}} \\
\midrule
\end{tabularx}

\vspace{0.6em}
\textbf{Perturbation (\(\mathcal{P}_y\): local paraphrase)}\\
\emph{Sunlight \textcolor{changeclr}{shimmers on a tranquil} pond. A fox hops near a lone tree.}

\vspace{0.35em}
\textbf{Perturbation (\(\mathcal{P}_y\): structural reorganization)}\\
\emph{\textcolor{changeclr}{A fox hops near a lone tree.} Sunlight glows on a still pond.}

\end{tcolorbox}

\caption{An example MCJudgeBench instance from the easy split, illustrating the instruction, candidate response, constraints, labels, and perturbation set of candidate response variants.}
\label{fig:mcjudge_example}
\end{figure}


    




\subsection{Benchmark Construction}
MCJudgeBench is built from two source benchmarks: ComplexBench \citep{wen2024complexbench} and InFoBench \citep{qin2024infobench}. We select these sources because they provide a progression in instruction length and constraint complexity. InFoBench contributes the \textit{easy} and \textit{hard} instances. The hard instances contain longer instructions and more decomposed requirements than the easy instances. ComplexBench contributes the \textit{complex} instances, which have substantially longer instructions than those in InFoBench and involve more complex multi-constraint structures. Together, these sources broaden the difficulty range covered by the benchmark, which consists of \textit{easy}, \textit{hard}, and \textit{complex} splits. We construct MCJudgeBench in three steps: source instance filtering, candidate construction, and human validation.

\textbf{Step 1: Source instance filtering.} We first preprocess source instances from ComplexBench and InFoBench using rule-based filtering to obtain a pool suitable for English-language constraint-level judge evaluation. To align the source data with our English-language evaluation setting, we apply text-level filters to the relevant instance fields, excluding cases with missing content or Chinese-language material. We additionally use metadata provided by the source benchmarks to exclude task types that tend to rely on culturally specific personas or background knowledge not contained in the prompt.

\textbf{Step 2: Candidate construction.} For each retained instance, we first generate a natural candidate response with Qwen3-4B-Instruct \citep{yang2025qwen3}. We use a relatively small open model because it produces responses with a more natural mix of satisfied, partially satisfied, and violated constraints than stronger frontier models, which are more likely to produce uniformly strong responses and therefore less informative benchmark instances. Human annotators then assign gold per-constraint adherence labels to the generated candidate response using the classes \textit{yes}, \textit{partial}, and \textit{no}. We also use GPT-5 mini to propose an initial single constraint-type label from the constraint text and to generate candidate response invariance perturbations. Prompt templates used in this pipeline are provided in Appendix~\ref{sec:appendix_method}.

\textbf{Step 3: Human validation.} Human annotators perform a final validation stage over the constructed instances. They review and correct the proposed constraint-type labels where needed, and may assign multiple final type labels when a constraint requires more than one form of verification. They also verify whether the generated perturbations preserve the annotated adherence label vector of the natural candidate response. We retain only perturbations that satisfy this criterion.

\paragraph{Constraint types.}
We group constraints in MCJudgeBench into eight types based on the verification challenge each constraint poses to an evaluator: \textit{Lexicon}, \textit{Numeric}, \textit{Format}, \textit{Content}, \textit{Component}, \textit{Faithfulness}, \textit{Factuality/Rationality}, and \textit{Style}. Our taxonomy is organized by the kind of judgment required to verify whether a constraint is satisfied. Full category definitions are provided in Appendix~\ref{sec:appendix_constraint_types}.

\paragraph{Invariance perturbation sets.}
We use response-side perturbations as invariance tests: the response form changes, but the per-constraint labels should remain unchanged \citep{ribeiro-etal-2020-beyond}. We generate candidate response variants from the natural responses using an LLM. Since many MCJudgeBench constraints directly depend on length, format, style, required keywords, required components, or factual content, broader transformations such as adding or removing text or changing style can alter whether a constraint is satisfied. We therefore restrict response-side perturbations to local changes in wording or structural organization.

\begin{itemize}[leftmargin=1.2em, itemsep=0pt, topsep=2pt, parsep=0pt]
    \item \textbf{Local paraphrase}: rewrites the response at the sentence or phrase level while preserving the same underlying content and adherence labels. It does not change required keywords, required structure, required components, or factual content relevant to the constraint labels.
    \item \textbf{Structural reorganization}: changes how the same content is arranged, for example by reordering independent sentences or regrouping content into different paragraphs. It does not change the required order, format, keywords, components, or any factual claim or judgment.
\end{itemize}

\paragraph{Inter-annotator agreement.}
Human validation is necessary because a meta-evaluation benchmark depends on reliable gold labels and validated invariance sets for assessing evaluator behavior. In our construction pipeline, annotators provide per-constraint adherence labels for the natural candidate responses and review subsequent artifacts, including proposed constraint types and candidate response perturbations.

To assess annotation consistency, three annotators independently label a shared subset comprising 10\% of instances from each split (\textit{easy}, \textit{hard}, and \textit{complex}). Table~\ref{tab:iaa} reports inter-annotator agreement on the adherence labels, using Fleiss' $\kappa$ for multi-rater agreement and Cohen's $\kappa$ for pairwise agreement. The agreement scores are consistently high across both measures, indicating strong annotation consistency.

\begin{table}[t]
\centering
\small
\begin{tabular}{lc}
\toprule
\textbf{Statistic} & \textbf{Value} \\
\midrule
Fleiss' $\kappa$ & 0.8768 \\
Cohen's $\kappa$ (A1--A2) & 0.8121 \\
Cohen's $\kappa$ (A1--A3) & 0.9286 \\
Cohen's $\kappa$ (A2--A3) & 0.8873 \\
\bottomrule
\end{tabular}
\caption{Inter-annotator agreement on the shared annotation subset. A1, A2, and A3 denote the three annotators.}
\label{tab:iaa}
\end{table}

\subsection{Benchmark Statistics}

\begin{table*}[t]
\centering
\small
\setlength{\tabcolsep}{4.0pt}
\renewcommand{\arraystretch}{1.08}
\begin{tabular}{lccccccccccc}
\toprule
\textbf{Split} & \textbf{\#Inst.} & \textbf{\#Con.} & \textbf{\#Con./Inst.} & \textbf{Len.} & \textbf{\#LP} & \textbf{\#SR} & \textbf{\#Pert.} & \textbf{\#Pert./Inst.} & \textbf{\#Yes} & \textbf{\#No} & \textbf{\#Partial} \\
\midrule
Easy    & 48  & 131 & 2.73 & 39.06  & 33 & 35 & 68  & 1.42 & 120 & 6  & 5  \\
Hard    & 52  & 336 & 6.46 & 60.06  & 42 & 32 & 74  & 1.42 & 293 & 31 & 12 \\
Complex & 41  & 186 & 4.54 & 228.46 & 32 & 27 & 59  & 1.44 & 142 & 28 & 16 \\
\midrule
Overall & 141 & 653 & 4.63 & 101.88 & 107 & 94 & 201 & 1.43 & 555 & 65 & 33 \\
\bottomrule
\end{tabular}
\caption{Summary statistics of MCJudgeBench, including the number of instructions (\#Inst.), total constraints (\#Con.), average constraints per instruction (\#Con./Inst.), average instruction length in words (Len.), response-side perturbations by type (local paraphrase: \#LP; structural reorganization: \#SR), total response-side perturbations (\#Pert.), average perturbations per instruction (\#Pert./Inst.), and the distribution of per-constraint gold labels.}
\label{tab:benchmark_stats}
\end{table*}

\begin{table}[t]
\centering
\small
\setlength{\tabcolsep}{4.5pt}
\renewcommand{\arraystretch}{1.08}
\begin{tabular}{lcc}
\toprule
\textbf{Constraint type} & \textbf{Count} & \textbf{\%} \\
\midrule
Component & 190 & 26.5 \\
Content & 118 & 16.5 \\
Factuality/Rationality & 99 & 13.8 \\
Faithfulness & 88 & 12.3 \\
Numeric & 76 & 10.6 \\
Format & 70 & 9.8 \\
Style & 51 & 7.1 \\
Lexicon & 25 & 3.5 \\
\bottomrule
\end{tabular}
\caption{Distribution of constraint types in MCJudgeBench. Of 653 constraints, 64 (9.8\%) have more than one type label. As a result, the type counts sum to more than the number of unique constraints.}

\label{tab:constraint_type_stats}
\end{table}

MCJudgeBench contains 141 instructions and 653 constraints. As shown in Table~\ref{tab:benchmark_stats}, the benchmark spans three splits with substantial differences in instruction length and constraint density. The distribution of gold labels also varies across splits, with complex instances containing a noticeably larger share of \textit{no} and \textit{partial} labels than easy and hard instances.

We generate two perturbation candidates per instance, for a total of 282 candidates. After human validation, 201 (71.3\%) are retained in the final benchmark. Most excluded candidates are due to infeasible perturbations for highly constrained outputs, where achieving a meaningful perturbation is difficult. Representative examples of excluded perturbation candidates are provided in Appendix~\ref{sec:appendix_rejected_perturbations}.

Table~\ref{tab:constraint_type_stats} summarizes the distribution of constraint types. Component and Content constraints are the most common, while Lexicon constraints are relatively rare. Some constraints receive more than one type label when they require multiple forms of verification, such as checking both required content and structured format.

\section{Baseline Evaluation}
\label{sec:baseline}

\subsection{Baselines}
We evaluate a mix of proprietary and open-source LLM judges to provide representative baselines for judge performance on MCJudgeBench. Our proprietary judges are GPT-5.2 \citep{openai_models}, Claude Sonnet 4.6 and Claude Haiku 4.5 \citep{anthropic_models}, and Gemini 3.1 Pro and Gemini 2.5 Flash-Lite \citep{google_gemini_models}. We also include two open-source judges, Qwen3.5-4B \citep{qwen2026qwen35_4b} and Llama 3.2 3B Instruct \citep{grattafiori2024llama32}, to complement the proprietary models with lightweight baselines under limited GPU resources. This selection provides a mix of stronger commercial judges, smaller cost-efficient variants, and lightweight open models that can be run under modest hardware constraints.

\subsection{Evaluation Protocol}
For each benchmark instance \(i\), a judge is given the instruction \(I_i\), a candidate response \(y_i\), and the explicit constraint list \(C_i=\{c_{i,1},\dots,c_{i,m_i}\}\). The judge predicts one label for each constraint from the set \(\{\textit{yes}, \textit{partial}, \textit{no}\}\), producing a label vector \(\hat{L}_i=\{\hat{\ell}_{i,1},\dots,\hat{\ell}_{i,m_i}\}\). These predictions are compared against the human-validated gold label vector \(L_i=\{\ell_{i,1},\dots,\ell_{i,m_i}\}\). We standardize the decoding temperature across models for each evaluation setting. For proprietary models, we evaluate both with and without reasoning when this setting is configurable. Open-source models are evaluated without reasoning as lightweight local baselines under limited GPU resources.

\paragraph{Measuring correctness.}
We first evaluate judge correctness on the original benchmark instances, using the natural candidate responses and a default evaluation prompt. To avoid introducing decoding randomness in this setting, we use deterministic decoding with \texttt{temperature}=0.0. Inspired by the Decomposed Requirements Following Ratio (DRFR) proposed in InFoBench \citep{qin2024infobench}, we aggregate correctness over individual constraints rather than relying only on instruction-level judgments. For instance \(i\), let
\[
a_{i,j}=\mathbf{1}\!\left[\hat{\ell}_{i,j}=\ell_{i,j}\right]
\]
denote whether the judge correctly predicts the gold label for constraint \(c_{i,j}\), where \(\ell_{i,j}\) is the human-validated gold label and \(\hat{\ell}_{i,j}\) is the corresponding judge prediction. We define the \textbf{Constraint-level Judge Accuracy Ratio (CJAR)} as
\[
\mathrm{CJAR}=\frac{\sum_{i=1}^{N}\sum_{j=1}^{m_i} a_{i,j}}{\sum_{i=1}^{N} m_i},
\]
where \(N\) is the number of benchmark instances and \(m_i\) is the number of constraints in instance \(i\). Because the gold labels are imbalanced across \textit{yes}, \textit{partial}, and \textit{no}, we also report macro-F1 over the three label classes as a complementary measure.

\paragraph{Measuring intrinsic inconsistency.}
To measure intrinsic inconsistency, we repeatedly query the same judge on the same instance under stochastic decoding while keeping the instruction, candidate response, constraint list, and evaluation prompt fixed. Let \(\hat{\ell}_{i,j}^{(1)},\dots,\hat{\ell}_{i,j}^{(K)}\) denote the predicted labels for constraint \(c_{i,j}\) across \(K\) repeated runs. In experiments, we set \(K=5\) and use stochastic decoding with \texttt{temperature}=1.0. We define the inconsistency indicator
\[
u_{i,j}=\mathbf{1}\!\left[\exists\, k\neq k' \text{ such that } \hat{\ell}_{i,j}^{(k)}\neq \hat{\ell}_{i,j}^{(k')}\right].
\]
The corresponding \textbf{intrinsic constraint-level inconsistency rate} is
\[
\mathrm{CIR}_{\text{intr}}=\frac{\sum_{i=1}^{N}\sum_{j=1}^{m_i} u_{i,j}}{\sum_{i=1}^{N} m_i}.
\]
This score measures the proportion of constraints whose predicted labels are not fully stable across repeated evaluations of the same underlying instance. As a more fine-grained companion measure, we also compute a pairwise disagreement version of intrinsic inconsistency, denoted \(\mathrm{CIR}_{\text{intr-pair}}\), whose full definition is provided in Appendix~\ref{sec:appendix_pairwise_intrinsic}.

\paragraph{Measuring procedural inconsistency.}
We evaluate judge stability under two invariance perturbation axes: evaluation prompt variants and candidate response variants. In both cases, the underlying gold label vector remains unchanged, so a stable judge should return the same constraint-level predictions across these alternative forms.

For evaluation prompt variants, we restate the same judging task while preserving the constraint content and required output schema. We consider three prompt variants: reordering the constraint list, changing the formatting of the constraints, and reordering the sections of the evaluation prompt template. Examples of the prompt variants are provided in Appendix~\ref{sec:appendix_eval_prompts}. For candidate response variants, we use the human-validated response perturbation sets described in Section~\ref{sec:benchmark}.

To isolate procedural effects from decoding randomness, we use deterministic decoding with \texttt{temperature}=0.0. For each instance \(i\), let \(V_i\) denote the set of valid prompt or response variants compared against the reference prediction. Let \(\hat{\ell}_{i,j}^{(0)}\) denote the reference prediction for constraint \(c_{i,j}\), and let \(\hat{\ell}_{i,j}^{(v)}\) denote the prediction under a prompt or response variant \(v \in V_i\). We define the pairwise inconsistency indicator
\[
u_{i,j}^{(v)}=\mathbf{1}\!\left[\hat{\ell}_{i,j}^{(v)}\neq \hat{\ell}_{i,j}^{(0)}\right].
\]
We then define
\[
\mathrm{CIR}_{\mathrm{proc}}
=
\frac{\sum_{i=1}^{N}\sum_{v\in V_i}\sum_{j=1}^{m_i} u_{i,j}^{(v)}}
{\sum_{i=1}^{N}\sum_{v\in V_i} m_i }.
\]

We report this quantity separately for prompt variants and response variants, yielding \(\mathrm{CIR}_{\text{prompt}}\) and \(\mathrm{CIR}_{\text{resp}}\), respectively. This metric is reference-relative: it counts any change from the reference prediction as inconsistency, regardless of whether the variant prediction is more or less correct with respect to the gold label. We therefore also include a gold-relative companion analysis of how often procedural perturbations change correctness with respect to the gold label. Full definitions are provided in Appendix~\ref{sec:appendix_proc_gold_relative}.

\subsection{Main Results}
We evaluate baseline judges on MCJudgeBench under three settings: correctness on the original benchmark instances, intrinsic inconsistency under repeated stochastic decoding, and procedural inconsistency under label-preserving invariance perturbations. The results are summarized in Table~\ref{tab:main_results}.

\definecolor{reasongray}{gray}{0.92}
\newcommand{\oncell}[1]{\cellcolor{reasongray}#1}

\begin{table*}[t]
\centering
\small
\setlength{\tabcolsep}{3.6pt}
\renewcommand{\arraystretch}{1.08}
\begin{tabular}{llcccccccc}
\toprule
\textbf{Model} &
\textbf{Reasoning} &
\multicolumn{2}{c}{\textbf{Correctness} $\uparrow$} &
\multicolumn{6}{c}{\textbf{Inconsistency (\%)} $\downarrow$} \\
\cmidrule(lr){3-4} \cmidrule(lr){5-10}
& &
\textbf{CJAR} &
\textbf{Macro-F1} &
\textbf{CIR$_{\text{intr}}$} &
\textbf{CIR$_{\text{intr-pair}}$} &
\textbf{CIR$_{\text{prompt}}$} &
\textbf{CIR$_{\text{prompt}}^{\dagger}$} &
\textbf{CIR$_{\text{resp}}$} &
\textbf{CIR$_{\text{resp}}^{\dagger}$} \\
\midrule
\multicolumn{10}{l}{\textit{Proprietary models}} \\

\multirow{2}{*}{GPT-5.2}
& Off & 0.775 & 0.592 & 9.49 & 4.49 & 9.04 & 9.04 & 7.51 & 7.51 \\
& \oncell{On} & \oncell{0.809} & \oncell{0.616} & \oncell{14.70} & \oncell{7.41} & \oncell{8.63} & \oncell{8.63} & \oncell{7.29} & \oncell{7.29} \\

\multirow{2}{*}{Claude Sonnet 4.6}
& Off & 0.821 & 0.607 & 5.72 & 2.61 & 6.36 & 6.74 & 4.97 & 4.97 \\
& \oncell{On} & \oncell{0.828} & \oncell{\underline{\textbf{0.637}}} & \oncell{5.21} & \oncell{2.51} & \oncell{6.18} & \oncell{6.18} & \oncell{4.02} & \oncell{4.02} \\

\multirow{2}{*}{Claude Haiku 4.5}
& Off & 0.822 & 0.585 & \underline{\textbf{2.94}} & \underline{\textbf{1.55}} & 5.66 & 6.33 & 4.48 & 5.29 \\
& \oncell{On} & \oncell{0.848} & \oncell{0.603} & \oncell{14.09} & \oncell{7.35} & \oncell{7.15} & \oncell{7.15} & \oncell{7.19} & \oncell{7.19} \\

\multirow{2}{*}{Gemini 2.5 Flash-Lite}
& Off & 0.855 & 0.518 & 7.35 & 3.61 & 5.64 & 6.13 & 3.81 & 3.81 \\
& \oncell{On} & \oncell{0.836} & \oncell{0.569} & \oncell{14.77} & \oncell{7.65} & \oncell{6.71} & \oncell{7.43} & \oncell{7.41} & \oncell{8.20} \\

Gemini 3.1 Pro
& \oncell{On} & \oncell{\underline{\textbf{0.858}}} & \oncell{0.598} & \oncell{6.58} & \oncell{3.32} & \oncell{\underline{\textbf{2.91}}} & \oncell{\underline{\textbf{2.91}}} & \oncell{4.02} & \oncell{4.02} \\

\midrule
\multicolumn{10}{l}{\textit{Open-source models}} \\

Qwen3.5-4B
& Off & 0.853 & 0.529 & 24.20 & 12.79 & 6.28 & 6.28 & \underline{\textbf{3.59}} & \underline{\textbf{3.59}} \\

Llama 3.2 3B Instruct
& N/A & 0.770 & 0.441 & 35.38 & 18.07 & 10.52 & 27.58 & 4.45 & 6.17 \\

\bottomrule
\end{tabular}

\caption{Main baseline results on MCJudgeBench. The Reasoning column indicates whether reasoning is enabled, with shaded cells denoting reasoning-enabled runs. N/A indicates that no reasoning mode is available. Correctness is measured by Constraint-level Judge Accuracy Ratio (CJAR) and macro-F1. Inconsistency is measured by constraint-level inconsistency rate (CIR). We report intrinsic inconsistency under repeated stochastic decoding (CIR$_{\text{intr}}$), its pairwise disagreement variant (CIR$_{\text{intr-pair}}$), procedural inconsistency under evaluation prompt variants (CIR$_{\text{prompt}}$), and procedural inconsistency under candidate response variants (CIR$_{\text{resp}}$). Columns marked with $\dagger$ apply parse-penalty scoring, where unparsable outputs are counted as inconsistent.}
\label{tab:main_results}
\end{table*}

\paragraph{Correctness.}
Claude Sonnet 4.6 with reasoning attains the highest macro-F1, while Gemini 3.1 Pro with reasoning achieves the highest CJAR. This contrast is meaningful because, in multi-constraint responses, most requirements are typically satisfied and only a smaller number are violated or partially satisfied. A judge can therefore obtain high CJAR by recognizing the majority of satisfied constraints while still missing rarer minority-label cases, especially \textit{no} and \textit{partial}, which are often more informative for reliability analysis. Qwen3.5-4B shows a similar pattern, remaining competitive on CJAR but achieving a lower macro-F1 than the stronger proprietary models. Its high CJAR should therefore not be interpreted as evidence of consistently strong performance across labels. The Qwen3.5-4B result may also be partly affected by the fact that Qwen3.5-4B is part of the same model family as the Qwen3-4B-Instruct model used for candidate response generation. Prior work on self-preference bias suggests that LLM judges can assign higher scores to outputs that are more likely under the judge model's own distribution, as reflected by lower perplexity \citep{wataoka2024self}. Across models evaluated with both reasoning settings, reasoning improves macro-F1, suggesting more balanced performance across label classes. By contrast, Llama 3.2 3B Instruct is noticeably weaker on both measures. Overall, these results reinforce the need to read CJAR together with macro-F1 and per-label behavior rather than treating CJAR rank alone as evidence of stronger judge quality.

\paragraph{Inconsistency.}
The clearest difference between proprietary and open-source models appears in intrinsic inconsistency, where the open-source baselines are much less stable under repeated stochastic decoding. This pattern is consistent under both the intrinsic metric and its pairwise variant. The effect of reasoning on inconsistency varies across inconsistency types and models. It reduces all inconsistency metrics for Claude Sonnet 4.6, increases them for Claude Haiku 4.5 and Gemini 2.5 Flash-Lite, and produces a mixed pattern for GPT-5.2, where intrinsic inconsistency increases but procedural inconsistency slightly decreases. Response-side procedural inconsistency is generally lower than intrinsic inconsistency, while prompt-side procedural inconsistency is more variable and is typically higher than response-side inconsistency. This suggests that judges' decisions are often more sensitive to changes in evaluation prompt form than to changes in candidate response form.

\begin{figure*}[t]
    \centering
    \includegraphics[width=\textwidth]{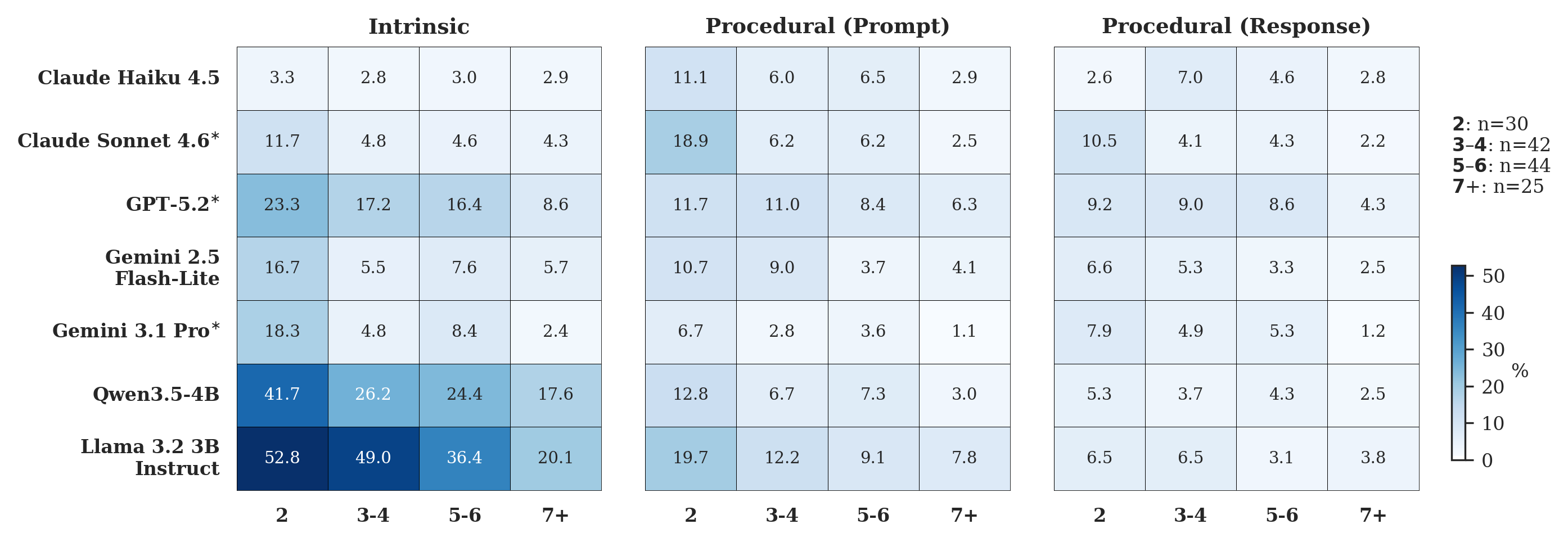}
    \caption{Constraint-level inconsistency rates by number of constraints. Models marked with $^{\ast}$ are evaluated with reasoning enabled. The intrinsic panel uses the primary intrinsic inconsistency metric, \(\mathrm{CIR}_{\text{intr}}\), while the other panels show procedural inconsistency under prompt-side and response-side perturbations.}
    \label{fig:inconsistency_by_num_constraints}
\end{figure*}

\begin{figure}[t]
    \centering
    \includegraphics[width=1.0\linewidth]{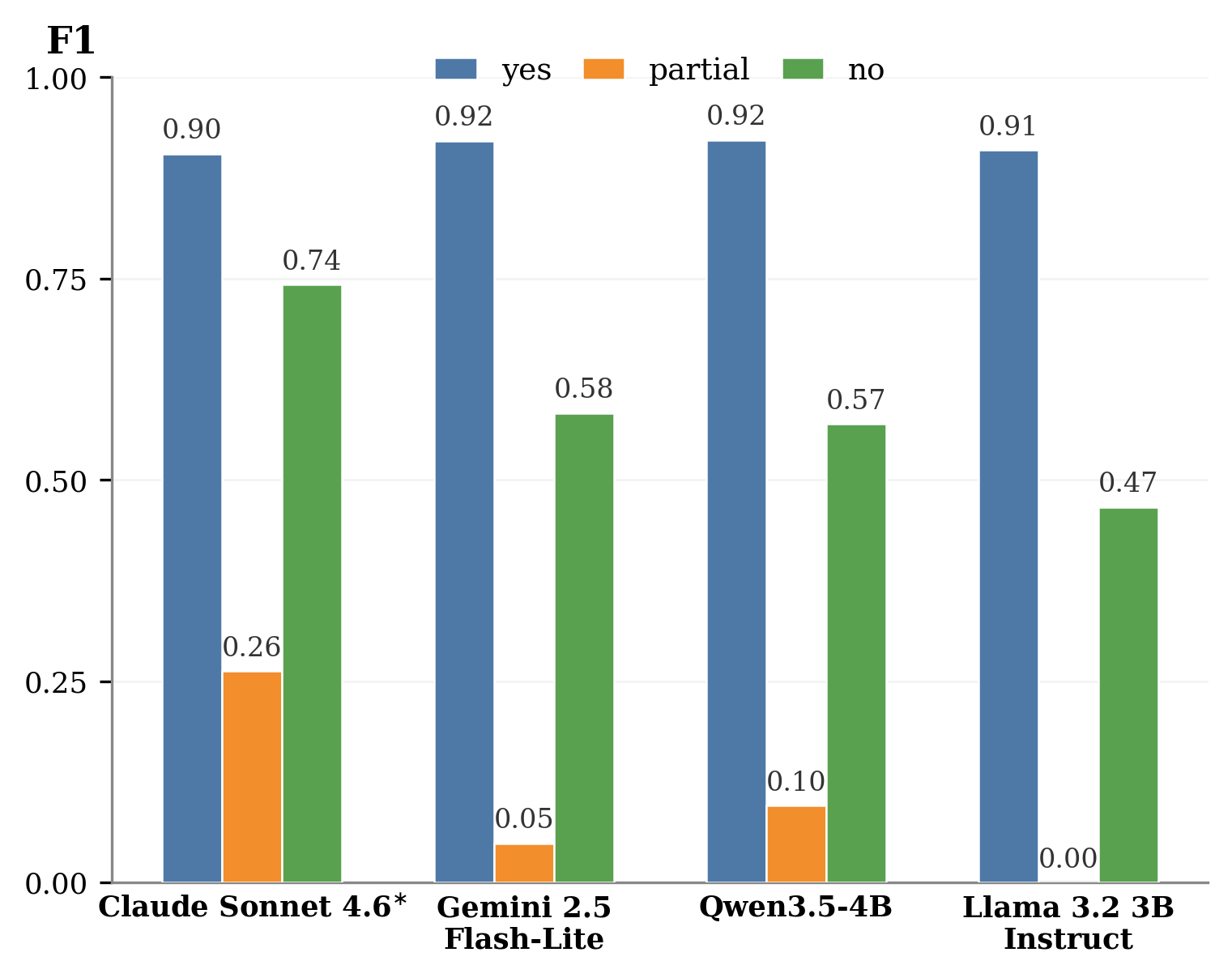}
    \caption{Per-label F1 for representative judges. Models marked with $^{\ast}$ are evaluated with reasoning enabled.}
    \label{fig:f1_by_label_representative}
\end{figure}

\paragraph{Interpretation.}
Taken together, these results suggest that judge reliability is not one-dimensional. Because multi-constraint responses usually satisfy most requirements and fail only a smaller number, high overall correctness can hide weaker detection of the minority but more informative \textit{no} and \textit{partial} cases. Reasoning also changes the reliability profile rather than uniformly improving it, since the gains in correctness do not always translate into lower inconsistency. This suggests that reasoning may change how judges resolve borderline constraints, improving some per-constraint predictions while also introducing additional variability across repeated or perturbed evaluations. This is broadly consistent with prior work showing that inference-time reasoning can be task-dependent rather than uniformly beneficial \citep{liu2024chain}. The large differences in intrinsic inconsistency, which remain visible under both the intrinsic metric and its pairwise variant, further suggest that some judges have less stable separation among \textit{yes}/\textit{partial}/\textit{no} verdicts, making small stochastic variation more likely to flip the predicted label. The gap between the standard and $\dagger$ procedural metrics also shows that, for some models, perturbations introduce another source of inconsistency beyond changed verdicts: the model may preserve the same underlying judgment but fail to return it in the required structured format. This makes format reliability, alongside verdict quality and verdict stability, an important part of judge reliability.

\subsection{Analysis}

We next examine where judge performance differs systematically across label categories, benchmark characteristics, and perturbation settings. These analyses help explain patterns that are not visible from aggregate correctness and inconsistency scores alone.

\paragraph{Performance by label.}
As shown in Figure~\ref{fig:f1_by_label_representative}, the gap between CJAR and macro-F1 is driven mainly by uneven performance across labels. Across representative judges, F1 is consistently high on \textit{yes}, lower on \textit{no}, and weakest by far on \textit{partial}. Thus, strong overall constraint-level accuracy does not imply equally strong detection of minority-label failures. Claude Sonnet 4.6 with reasoning is the most balanced of the representative judges, whereas Gemini 2.5 Flash-Lite, Qwen3.5-4B, and Llama 3.2 3B Instruct remain strongest on \textit{yes} and substantially weaker on \textit{partial}. This explains why some models remain competitive on CJAR while underperforming on macro-F1. The confusion matrices in Appendix~\ref{sec:appendix_confusion} show that these errors are often asymmetric: gold \textit{partial} cases are more often mapped to \textit{yes} than to \textit{partial}, indicating a tendency to over-credit partially satisfied constraints rather than distinguish them reliably as a separate adherence state.

\paragraph{Patterns of inconsistency.}
Figure~\ref{fig:inconsistency_by_num_constraints} shows that inconsistency does not increase monotonically with constraint count. For most models, intrinsic inconsistency is highest on 2-constraint instructions and declines as more constraints are added, although Claude Haiku 4.5 is a notable exception with uniformly low intrinsic inconsistency across all constraint counts. Prompt-side procedural inconsistency shows a similar tendency, while response-side procedural inconsistency is lower overall and varies less with constraint count. This is counter-intuitive because instructions with more constraints would naturally seem harder to judge. One possible explanation is that constraint count does not directly measure judgment ambiguity. Longer constraint lists may make the verification task more explicitly decomposed, whereas 2-constraint instructions may leave the boundary between \textit{yes}, \textit{partial}, and \textit{no} less distinct, making predictions more sensitive to stochastic variation and procedural reformulation. This suggests that judge instability is not determined by constraint count alone. Additional analyses by constraint type are provided in Appendix~\ref{sec:appendix_inconsistency_types}.

\paragraph{Implications for judge reliability.}
Correctness and inconsistency capture related but distinct aspects of judge reliability. The model rankings in Table~\ref{tab:main_results} do not reduce to a single ordering: Claude Sonnet 4.6 with reasoning achieves the strongest macro-F1, but not the lowest inconsistency on every axis, while Claude Haiku 4.5 without reasoning shows particularly low intrinsic inconsistency without corresponding gains in macro-F1. Appendix~\ref{sec:appendix_proc_gold_relative} further shows that reference-relative procedural inconsistency should not be interpreted as a direct proxy for correctness degradation: when procedural perturbations change correctness, the distribution of transitions is often fairly balanced between correct-to-incorrect and incorrect-to-correct cases. Together, these findings reinforce the need to evaluate both correctness and stability rather than treating either as a sufficient proxy for overall judge reliability.

\section{Conclusion}
We introduced \textbf{MCJudgeBench}, a benchmark for constraint-level judge evaluation in multi-constraint instruction following. MCJudgeBench combines explicit constraint lists, per-constraint gold labels, and controlled response-side invariance sets, together with an evaluation protocol that includes prompt variants, enabling direct study of whether an evaluator can correctly and stably verify individual requirements.

Our results show that judge reliability is not captured by a single score. High overall correctness can hide weak detection of \textit{partial} and \textit{no} cases, and higher correctness does not necessarily imply stronger stability under repeated or perturbed evaluation. The effect of reasoning varies across models and evaluation metrics, rather than uniformly improving judge reliability. In particular, intrinsic inconsistency under stochastic decoding emerges as an important source of unreliability, while procedural inconsistency reveals additional sensitivity to prompt and response reformulation, with perturbations sometimes degrading judgments and sometimes correcting them.

These findings suggest that evaluating LLM judges for multi-constraint instruction following requires joint consideration of correctness and stability at the constraint level. We hope MCJudgeBench provides a useful benchmark for future work on more reliable and robust automated evaluators.

\section*{Limitations}
MCJudgeBench focuses on English-language multi-constraint instruction following and is constructed from two source benchmarks, ComplexBench and InFoBench. Its coverage therefore does not extend to all instruction following task distributions or languages. Although we include controlled perturbations over the candidate response and evaluation prompt forms, these do not exhaust all possible sources of judge instability. In addition, although human validation improves label quality, distinguishing \textit{partial} from \textit{yes} or \textit{no} can still be difficult in borderline cases. Some constraints may also admit multiple reasonable interpretations, leaving residual annotation subjectivity despite strong inter-annotator agreement.

\section*{Ethics}
\paragraph{Data sources and licensing.}
MCJudgeBench is constructed from two publicly released benchmark sources, ComplexBench and InFoBench, which are available under permissive licenses for research use, and is further extended through additional human validation.

\paragraph{Human subjects.}
The human annotation stage was conducted for research benchmark construction and quality control. The annotation task involved judging constraint adherence and validating label-preserving perturbations, and did not require the collection of personal or sensitive information from annotators or from the benchmark content.



\bibliography{custom}

\appendix

\section{Constraint Type Definitions}
\label{sec:appendix_constraint_types}

Below, we give the full definitions of the eight constraint types used in MCJudgeBench.

\begin{itemize}[leftmargin=1.2em, itemsep=2pt, topsep=2pt, parsep=0pt]
    \item \textbf{Lexicon}: checks whether specific words, phrases, or symbols appear in the output.
    \item \textbf{Numeric}: checks explicit counts or bounds, such as the number of words, characters, sentences, or items.
    \item \textbf{Format}: checks whether the output follows a required structure or layout, such as JSON, Markdown, tables, headings, bullet lists, or fixed templates.
    \item \textbf{Content}: checks whether the response is of the required high-level kind, such as a summary, report, itinerary, or tutorial.
    \item \textbf{Component}: checks whether specific information elements or sections are included in the response.
    \item \textbf{Faithfulness}: checks whether the response is grounded in the provided input or reference, including cases where correctness is fully determined by the given context.
    \item \textbf{Factuality/Rationality}: checks correctness or plausibility beyond the provided input, including logical or plausibility judgments based on internal knowledge.
    \item \textbf{Style}: checks whether the response matches a required tone, voice, register, or persona, such as being formal, polite, concise, or beginner-friendly.
\end{itemize}

\section{Prompt Templates for Benchmark Construction and Evaluation}
\label{sec:appendix_method}

\subsection{Natural Candidate Response Generation}
\label{sec:appendix_candidate_generation}

To construct natural candidate responses, we use a simple instruction following prompt that asks the model to answer the source instruction directly and return only the response itself.

\begin{lstlisting}[style=promptstyle]
Your task is to produce a response to the instruction below.

Output ONLY the final response content.

Instruction:
{instruction}

Input:
{input}

Response:
\end{lstlisting}
When no additional input field is available, the \texttt{\{input\}} field is omitted.

\subsection{Constraint Type Proposal}
\label{sec:appendix_constraint_type_proposal}
We use an LLM to propose an initial single constraint type for each constraint based on the taxonomy described in Section~\ref{sec:benchmark}. This step only assists benchmark construction. During human validation, annotators review and revise the proposals, and may assign multiple final type labels if a constraint spans more than one verification category.

\begin{lstlisting}[style=promptstyle]
Assign the input constraint text to ONE category from the taxonomy below.

TASK
Given a single constraint sentence, return the most appropriate category label.
- Output format MUST be valid JSON:
  {"const_type": ["Category"]}

ALLOWED CATEGORY LABELS
- "Lexicon"
- "Numeric"
- "Format"
- "Content"
- "Component"
- "Faithfulness"
- "Factuality/Rationality"
- "Style"

TAXONOMY

Lexicon
Checks whether specific words, phrases, or symbols appear in the output.

Numeric
Checks explicit counts or bounds, such as the number of words, characters, sentences, or items.

Format
Checks whether the output follows a required structure or layout, such as JSON, Markdown, tables, headings, bullet lists, or fixed templates.

Content
Checks whether the response is of the required high-level kind, such as a summary, report, itinerary, or tutorial.

Component
Checks whether specific information elements or sections are included in the response.

Faithfulness
Checks whether the response is grounded in the provided input or reference, including cases where correctness is fully determined by the given context.

Factuality/Rationality
Checks correctness or plausibility beyond the provided input, including logical or plausibility judgments based on internal knowledge.

Style
Checks whether the response matches a required tone, voice, register, or persona, such as being formal, polite, concise, or beginner-friendly.

INPUT
Constraint: {constraint}

OUTPUT
Return ONLY the JSON object.
\end{lstlisting}

\subsection{Candidate Response Perturbation Generation}
\label{sec:appendix_response_perturbation_generation}
We use an LLM to generate candidate response perturbations that preserve the original constraint adherence labels. The perturbation prompt conditions on the instruction, the response, and the partition of constraints into \textit{yes}, \textit{partial}, and \textit{no}. The model is instructed to rewrite the response while preserving the same label for each constraint. We use the following template for two perturbation types.

\begin{lstlisting}[style=promptstyle]
You are generating a candidate response invariance perturbation for a benchmark instance.

Your goal is to rewrite the original response while preserving the exact same constraint adherence labels.

TASK TYPE:
{perturbation_type}

INSTANCE
Instruction:
{instruction}

Original response:
{output}

CONSTRAINTS THAT ARE SATISFIED (label = yes)
These constraints are satisfied in the original response and must remain satisfied after rewriting.
{yes_constraints}

CONSTRAINTS THAT ARE PARTIALLY SATISFIED (label = partial)
These constraints are only partially satisfied in the original response and must remain partially satisfied after rewriting.
{partial_constraints}

CONSTRAINTS THAT ARE VIOLATED (label = no)
These constraints are violated in the original response and must remain violated after rewriting.
{no_constraints}

REQUIREMENTS
1. Preserve the exact same adherence label for every constraint.
2. Do not add, remove, or change anything that would alter a constraint label.
3. If no safe rewrite is possible, output NO_SAFE_PERTURBATION.

Return only the rewritten response, or NO_SAFE_PERTURBATION.
\end{lstlisting}

\paragraph{Local paraphrase.}
This variant allows only local wording changes, such as minor rephrasing, sentence-level paraphrases, discourse marker changes, or local simplification and compression. It must not change the required keywords or phrases, required format or section structure, the presence or absence of required components, factual claims, or any wording whose change would alter a constraint label. It also must not repair violated constraints or complete partially satisfied ones.

\paragraph{Structural reorganization.}
This variant allows only organizational changes, such as reordering independent sentences or sections, regrouping content into different paragraphs, or reordering independent bullet items when order is not required. It must not change required order or sequencing, required prefix or suffix position, required format type, required keywords, the presence or absence of required components, or any factual claim, verdict, or reasoning content. It also must not make the response more complete or less complete in a way that would alter any constraint label.

\subsection{Evaluation Prompt Templates and Variants}
\label{sec:appendix_eval_prompts}

All baseline judges are evaluated with a base prompt that asks the model to verify each constraint independently and assign one label from \textit{yes}, \textit{partial}, and \textit{no}.

\begin{lstlisting}[style=promptstyle]
You are a strict evaluator of instruction following constraints.

You will be given:
- an instruction
- a model output
- a list of constraints

For each constraint, assign exactly one label:
- yes: clearly satisfied
- no: clearly violated or missing
- partial: partially satisfied but not fully satisfied

Rules:
- Judge each constraint independently.
- Use only the provided instruction, model output, and constraint text.
- Do not assume missing information.
- Do not reward fluency, style, or overall quality if a constraint is violated.
- Use partial only when there is clear evidence that the constraint is satisfied in part but not in full.

INSTRUCTION:
{instruction}

MODEL OUTPUT:
{output}

CONSTRAINTS:
{constraints}

Return ONLY valid JSON with EXACTLY this schema:
{output_format}
\end{lstlisting}
\paragraph{Prompt variants.}
To measure prompt-side procedural inconsistency, we construct semantically equivalent variants of the base evaluation prompt. These variants preserve the instruction, candidate response, constraint content, and required output schema, while changing only the presentation of the evaluation task. Below, we illustrate the three prompt perturbation types used in our experiments.

\paragraph{Constraint reordering.}
The constraint list is permuted, while the instruction, model output, constraint text, and output schema remain unchanged. A default constraint order
\begin{lstlisting}[style=promptstyle]
CONSTRAINTS:
1. {constraint_1}
2. {constraint_2}
3. {constraint_3}
\end{lstlisting}
is changed to
\begin{lstlisting}[style=promptstyle]
CONSTRAINTS:
3. {constraint_3}
2. {constraint_2}
1. {constraint_1}
\end{lstlisting}

\paragraph{Constraint formatting.}
The constraint content is unchanged, but the presentation format of the constraint list is altered. A numbered list
\begin{lstlisting}[style=promptstyle]
CONSTRAINTS:
1. {constraint_1}
2. {constraint_2}
3. {constraint_3}
\end{lstlisting}
is rewritten as a structured field format:
\begin{lstlisting}[style=promptstyle]
CONSTRAINTS:
- constraint_id: "1"
  text: "{constraint_1}"
- constraint_id: "2"
  text: "{constraint_2}"
- constraint_id: "3"
  text: "{constraint_3}"
\end{lstlisting}

\paragraph{Template reordering.}
The same prompt sections are preserved, but their order is changed. The default section order
\begin{lstlisting}[style=promptstyle]
INSTRUCTION:
{instruction}

MODEL OUTPUT:
{output}

CONSTRAINTS:
{constraints}
\end{lstlisting}
is changed to
\begin{lstlisting}[style=promptstyle]
CONSTRAINTS:
{constraints}

INSTRUCTION:
{instruction}

MODEL OUTPUT:
{output}
\end{lstlisting}

\section{Human Annotation and Validation}
\label{sec:appendix_annotation}

\subsection{Annotation Criteria}
\label{sec:appendix_annotation_criteria}

For each explicit constraint, annotators assign one label from \textit{yes}, \textit{partial}, and \textit{no}. Here, \textit{yes} denotes clear satisfaction, \textit{no} denotes clear violation or omission, and \textit{partial} is used only when the constraint is met in part with clear evidence. Annotations are made independently for each constraint using only the instruction, candidate response, and constraint text. Annotators do not assume missing information or reward overall fluency or response quality when a specific constraint is not met.

For candidate response perturbations, annotators verify that the rewritten response preserves the per-constraint label vector. Perturbations are retained only when all constraint labels remain unchanged.

\section{Excluded Perturbation Examples}
\label{sec:appendix_rejected_perturbations}

\subsection{Infeasible Perturbations}
Figure~\ref{fig:excluded_perturbation_infeasible} illustrates this failure mode for a highly constrained structured output, while Figure~\ref{fig:excluded_perturbation_minimal} shows the same issue for a minimal output.

\definecolor{yesgreen}{RGB}{0,120,60}
\definecolor{nored}{RGB}{170,40,40}
\begin{figure}[t]
\centering
\footnotesize

\begin{tcolorbox}[
    colback=gray!4,
    colframe=black!55,
    boxrule=0.5pt,
    arc=1mm,
    left=1mm,right=1mm,top=1mm,bottom=1mm,
    width=\columnwidth
]
\textbf{Instruction (\(I\), excerpt)}\\
Create a floor plan for a 2-bedroom, 2-bathroom apartment with a living room and a kitchen. Use \texttt{"\_"} and \texttt{"|"} to represent walls, include room labels, and ensure that the outer walls enclose all text and interior walls.

\vspace{0.6em}
\textbf{Candidate response (\(y\))}
\begin{lstlisting}[style=asciistyle]
_______________________________
|   Bedroom       |   Living Room
|____________     |
|   Bathroom      |
________________________________
\end{lstlisting}

\vspace{0.3em}

\renewcommand{\arraystretch}{1.08}
\setlength{\tabcolsep}{4pt}
\begin{tabularx}{\linewidth}{@{}>{\raggedright\arraybackslash}X >{\raggedleft\arraybackslash}p{0.24\linewidth}@{}}
\textbf{Constraint \(c_j \in C\)} & \textbf{Label \(\ell_j \in L\)} \\
\midrule
1. Does the generated text represent a floor plan?
& \textit{\textcolor{yesgreen}{yes}} \\

2. Are the symbols \texttt{"\_"} and \texttt{"|"} used to represent walls?
& \textit{\textcolor{yesgreen}{yes}} \\

3. Are text labels used to indicate room functions?
& \textit{\textcolor{yesgreen}{yes}} \\

4. Are all text labels and interior walls fully enclosed by outer walls?
& \textit{\textcolor{nored}{no}} \\

5. Does the generated text depict a 2-bedroom, 2-bathroom apartment with a living room and a kitchen?
& \textit{\textcolor{nored}{no}} \\
\midrule
\end{tabularx}

\vspace{0.6em}
\textbf{Reason for exclusion}\\
A label-preserving perturbation is infeasible here because the floor-plan layout is highly constrained. Meaningful changes would risk altering the wall structure, room layout, or room labels.
\end{tcolorbox}
\caption{Infeasible perturbation for a structured output.}
\label{fig:excluded_perturbation_infeasible}
\end{figure}

\begin{figure}[t]
\centering
\footnotesize

\begin{tcolorbox}[
    colback=gray!4,
    colframe=black!55,
    boxrule=0.5pt,
    arc=1mm,
    left=1mm,right=1mm,top=1mm,bottom=1mm,
    width=\columnwidth
]

\textbf{Instruction (\(I\), excerpt)}\\
Provide an ESRB rating for the following game. Then give an explanation of no more than 50 characters without any emotional attitude. Determine if the game content is R-rated before giving a rating:
\begin{itemize}[leftmargin=1.2em, itemsep=1pt, topsep=2pt, parsep=0pt]
    \item If it is, output \texttt{R-rated};
    \item If not, directly give the rating.
\end{itemize}

\vspace{-1em}

\begin{quote}
This is an action-adventure game where players help Kratos and his son complete a dangerous mission. Players explore various realms and engage in intense melee combat ... often resulting in large blood splatter effects and dismemberment ... repeatedly chopping a creature's neck with an axe results in decapitation. Words like \texttt{"f**k"} and \texttt{"sh*t"} can be heard in the game.
\end{quote}

\textbf{Candidate response (\(y\))}\\
\emph{R-rated}

\vspace{0.6em}
\renewcommand{\arraystretch}{1.08}
\setlength{\tabcolsep}{4pt}
\begin{tabularx}{\linewidth}{@{}>{\raggedright\arraybackslash}X >{\raggedleft\arraybackslash}p{0.24\linewidth}@{}}
\textbf{Constraint \(c_j \in C\)} & \textbf{Label \(\ell_j \in L\)} \\
\midrule
1. Does the model correctly determine the game content to be R-rated?
& \textit{\textcolor{yesgreen}{yes}} \\

2. Does the model respond with \texttt{R-rated}?
& \textit{\textcolor{yesgreen}{yes}} \\

3. Does the model provide an explanatory text for the game being classified as \texttt{R-rated} after responding with \texttt{R-rated}?
& \textit{\textcolor{nored}{no}} \\

4. Does the number of characters in the explanatory text output by the model not exceed 50 characters?
& \textit{\textcolor{nored}{no}} \\

5. Does the model output explanation without any emotional attitude?
& \textit{\textcolor{nored}{no}} \\
\midrule
\end{tabularx}

\vspace{0.6em}
\textbf{Reason for exclusion}\\
A label-preserving perturbation is infeasible here because the response must remain exactly \texttt{R-rated}.
\end{tcolorbox}
\caption{Infeasible perturbation for a minimal output.}
\label{fig:excluded_perturbation_minimal}
\end{figure}

\section{Detailed Quantitative Analysis}
\label{sec:appendix_quantitative}

\subsection{Pairwise Intrinsic Inconsistency}
\label{sec:appendix_pairwise_intrinsic}
In addition to the intrinsic inconsistency rate, we compute a more fine-grained pairwise disagreement version that measures the extent of disagreement across repeated stochastic evaluations. For each constraint \(c_{i,j}\), let \(\hat{\ell}_{i,j}^{(1)},\dots,\hat{\ell}_{i,j}^{(K)}\) denote the predicted labels across \(K\) repeated runs. We define the per-constraint pairwise disagreement score as
\[
d_{i,j}
=
\frac{1}{\binom{K}{2}}
\sum_{1 \le k < k' \le K}
\mathbf{1}\!\left[\hat{\ell}_{i,j}^{(k)}\neq \hat{\ell}_{i,j}^{(k')}\right],
\]
which is the fraction of run pairs that disagree for that constraint. We then define the benchmark-level pairwise intrinsic inconsistency as
\[
\mathrm{CIR}_{\text{intr-pair}}
=
\frac{\sum_{i=1}^{N}\sum_{j=1}^{m_i} d_{i,j}}{\sum_{i=1}^{N} m_i}.
\]
Unlike the standard metric, which records only whether any disagreement occurs, this companion metric captures the degree of disagreement across repeated runs.

\subsection{Gold-Relative Procedural Analysis}
\label{sec:appendix_proc_gold_relative}

To complement the reference-relative procedural inconsistency metric, we additionally analyze procedural perturbations with respect to the gold label. This companion analysis measures how often a procedural perturbation changes correctness relative to the gold label, and then decomposes those cases into transitions from correct to incorrect and from incorrect to correct.

For each constraint \(c_{i,j}\), where \(m_i\) is the number of constraints in instance \(i\), let \(\hat{\ell}_{i,j}^{(0)}\) denote the reference prediction on the original instance, let \(\hat{\ell}_{i,j}^{(v)}\) denote the prediction under a prompt or response variant \(v \in V_i\), and let \(\ell_{i,j}\) denote the gold label. We define the correctness indicators
\[
r_{i,j}^{(0)}=\mathbf{1}\!\left[\hat{\ell}_{i,j}^{(0)}=\ell_{i,j}\right]
\quad\text{and}\quad
r_{i,j}^{(v)}=\mathbf{1}\!\left[\hat{\ell}_{i,j}^{(v)}=\ell_{i,j}\right].
\]

We define the correctness-change rate as
\[
R_{\Delta\mathrm{corr}}
=
\frac{
\sum_{i=1}^{N}\sum_{v\in V_i}\sum_{j=1}^{m_i}
\mathbf{1}\!\left[r_{i,j}^{(0)}\neq r_{i,j}^{(v)}\right]
}{
\sum_{i=1}^{N}\sum_{v\in V_i} m_i
}.
\]

Among the cases where correctness changes, we define the correct-to-incorrect proportion as
\[
P_{\text{c}\to\text{i}}
=
\frac{
\sum\limits_{i=1}^{N}\sum\limits_{v\in V_i}\sum\limits_{j=1}^{m_i}
\mathbf{1}\!\left[r_{i,j}^{(0)}=1 \land r_{i,j}^{(v)}=0\right]
}{
\sum\limits_{i=1}^{N}\sum\limits_{v\in V_i}\sum\limits_{j=1}^{m_i}
\mathbf{1}\!\left[r_{i,j}^{(0)}\neq r_{i,j}^{(v)}\right]
},
\]
and the incorrect-to-correct proportion as
\[
P_{\text{i}\to\text{c}}
=
\frac{
\sum\limits_{i=1}^{N}\sum\limits_{v\in V_i}\sum\limits_{j=1}^{m_i}
\mathbf{1}\!\left[r_{i,j}^{(0)}=0 \land r_{i,j}^{(v)}=1\right]
}{
\sum\limits_{i=1}^{N}\sum\limits_{v\in V_i}\sum\limits_{j=1}^{m_i}
\mathbf{1}\!\left[r_{i,j}^{(0)}\neq r_{i,j}^{(v)}\right]
}.
\]

By construction, \(P_{\text{c}\to\text{i}} + P_{\text{i}\to\text{c}} = 1\) whenever \(R_{\Delta\mathrm{corr}} > 0\).

\begin{table*}[t]
\centering
\small
\setlength{\tabcolsep}{3.6pt}
\renewcommand{\arraystretch}{1.08}
\begin{tabular}{llcccccc}
\toprule
\textbf{Model} &
\textbf{Reasoning} &
\multicolumn{3}{c}{\textbf{Prompt-side (\%)}} &
\multicolumn{3}{c}{\textbf{Response-side (\%)}} \\
\cmidrule(lr){3-5} \cmidrule(lr){6-8}
& &
\(\mathbf{R_{\Delta\mathrm{corr}}}\) &
\(\mathbf{P_{\text{c}\to\text{i}}}\) &
\(\mathbf{P_{\text{i}\to\text{c}}}\) &
\(\mathbf{R_{\Delta\mathrm{corr}}}\) &
\(\mathbf{P_{\text{c}\to\text{i}}}\) &
\(\mathbf{P_{\text{i}\to\text{c}}}\) \\
\midrule
\multicolumn{8}{l}{\textit{Proprietary models}} \\

\multirow{2}{*}{GPT-5.2}
& Off & 7.10 & 64.03 & 35.97 & 6.24 & 49.15 & 50.85 \\
& \oncell{On} & \oncell{6.02} & \oncell{58.47} & \oncell{41.53} & \oncell{4.86} & \oncell{41.30} & \oncell{58.70} \\

\multirow{2}{*}{Claude Sonnet 4.6}
& Off & 4.61 & 50.00 & 50.00 & 3.59 & 50.00 & 50.00 \\
& \oncell{On} & \oncell{4.70} & \oncell{47.83} & \oncell{52.17} & \oncell{3.49} & \oncell{51.52} & \oncell{48.48} \\

\multirow{2}{*}{Claude Haiku 4.5}
& Off & 3.91 & 53.95 & 46.05 & 3.30 & 58.06 & 41.94 \\
& \oncell{On} & \oncell{5.16} & \oncell{56.44} & \oncell{43.56} & \oncell{5.07} & \oncell{56.25} & \oncell{43.75} \\

\multirow{2}{*}{Gemini 2.5 Flash-Lite}
& Off & 4.72 & 47.83 & 52.17 & 3.38 & 37.50 & 62.50 \\
& \oncell{On} & \oncell{5.56} & \oncell{57.01} & \oncell{42.99} & \oncell{6.02} & \oncell{53.57} & \oncell{46.43} \\

Gemini 3.1 Pro
& \oncell{On} & \oncell{2.65} & \oncell{46.15} & \oncell{53.85} & \oncell{3.70} & \oncell{31.43} & \oncell{68.57} \\

\midrule
\multicolumn{8}{l}{\textit{Open-source models}} \\

Qwen3.5-4B
& Off & 5.56 & 42.20 & 57.80 & 2.96 & 53.57 & 46.43 \\

Llama 3.2 3B Instruct
& N/A & 9.56 & 51.43 & 48.57 & 4.00 & 60.00 & 40.00 \\

\bottomrule
\end{tabular}

\caption{Gold-relative procedural analysis by model. \(R_{\Delta\mathrm{corr}}\) measures how often procedural perturbations change correctness relative to the gold label, while \(P_{\text{c}\to\text{i}}\) and \(P_{\text{i}\to\text{c}}\) decompose those correctness-changing cases by direction. Shaded cells denote reasoning-enabled runs.}
\label{tab:gold_relative_proc}
\end{table*}

Table~\ref{tab:gold_relative_proc} shows that prompt-side perturbations generally change correctness more often than response-side perturbations. This is consistent with the earlier pattern that \(\mathrm{CIR}_{\text{prompt}}\) is typically higher than \(\mathrm{CIR}_{\text{resp}}\). Reasoning-enabled runs support the same qualitative point, since reasoning can change the rate and direction of correctness changes, but procedural perturbations are not uniformly harmful or uniformly beneficial. When correctness changes, the direction of change is often relatively balanced across models, with correct-to-incorrect and incorrect-to-correct transitions mostly falling within a roughly 40--60\% range, although some settings are more skewed. This indicates that reference-relative procedural inconsistency should not be interpreted as a direct proxy for correctness degradation. Some perturbation-induced changes make the prediction less correct, while others make it more correct.

\subsection{Confusion Patterns Across Models}
\label{sec:appendix_confusion}

\begin{figure*}[t]
    \centering
    \includegraphics[width=\textwidth]{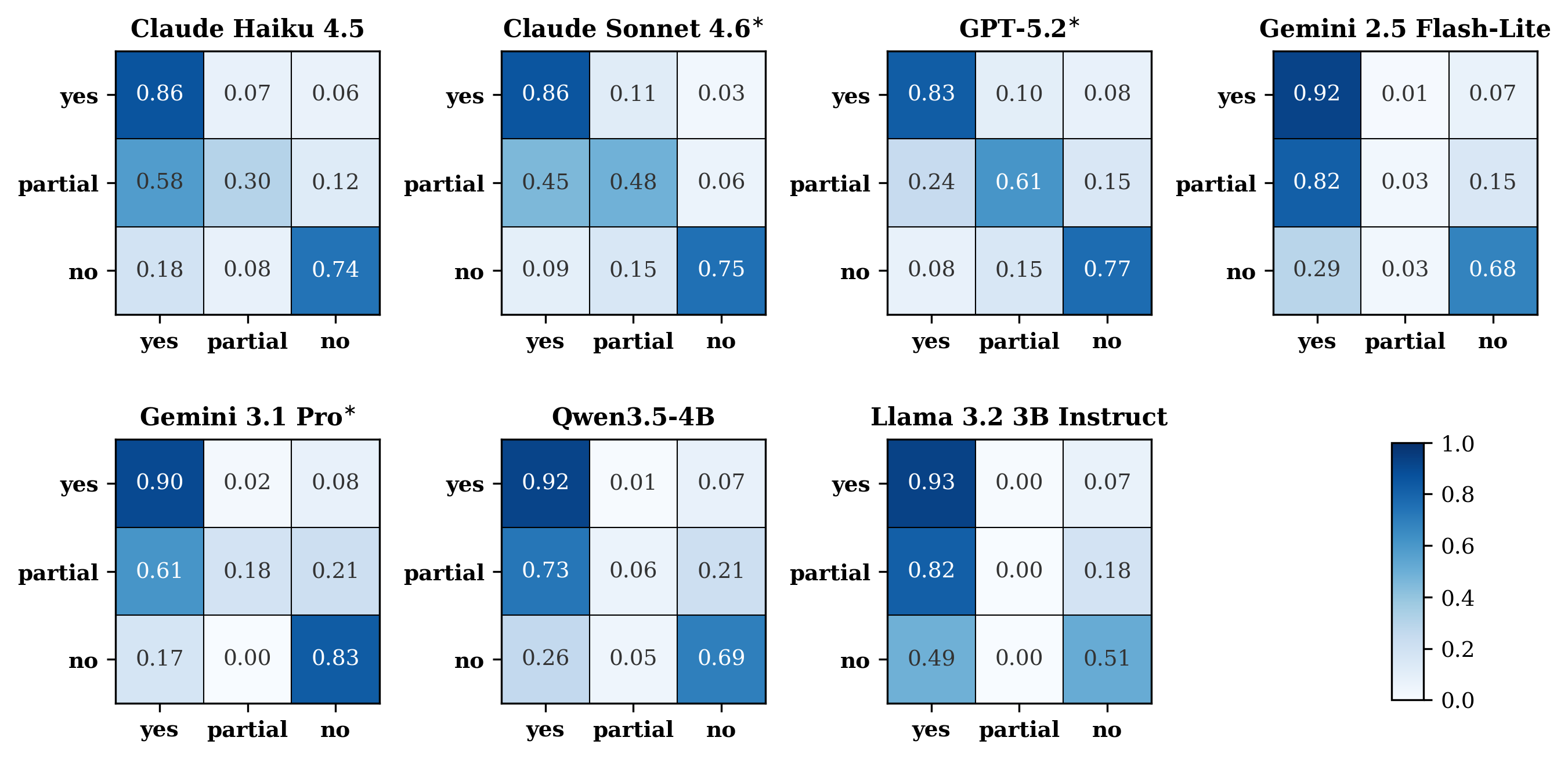}
    \caption{Confusion matrices for all baseline judges, with each gold-label row normalized to sum to one. Rows correspond to gold labels and columns to predicted labels. Models marked with $^{\ast}$ are evaluated with reasoning enabled.}
    \label{fig:confusion_grid_all_models}
\end{figure*}

Figure~\ref{fig:confusion_grid_all_models} shows confusion matrices for all baseline judges, with each gold-label row normalized to sum to one. These matrices complement the per-label F1 plot for representative judges in the main text by showing how each gold label is distributed across predicted labels. Across models, the weakest label is typically \textit{partial}. For most judges, gold \textit{partial} cases are more often mapped to \textit{yes} than to \textit{partial}, indicating a tendency to over-credit partially satisfied constraints rather than distinguish them reliably as a separate adherence state. By contrast, gold \textit{yes} cases are identified much more reliably, while performance on gold \textit{no} is generally intermediate between these two extremes.

\subsection{Inconsistency Across Constraint Types}
\label{sec:appendix_inconsistency_types}

\begin{figure*}[t]
    \centering
    \includegraphics[width=\textwidth]{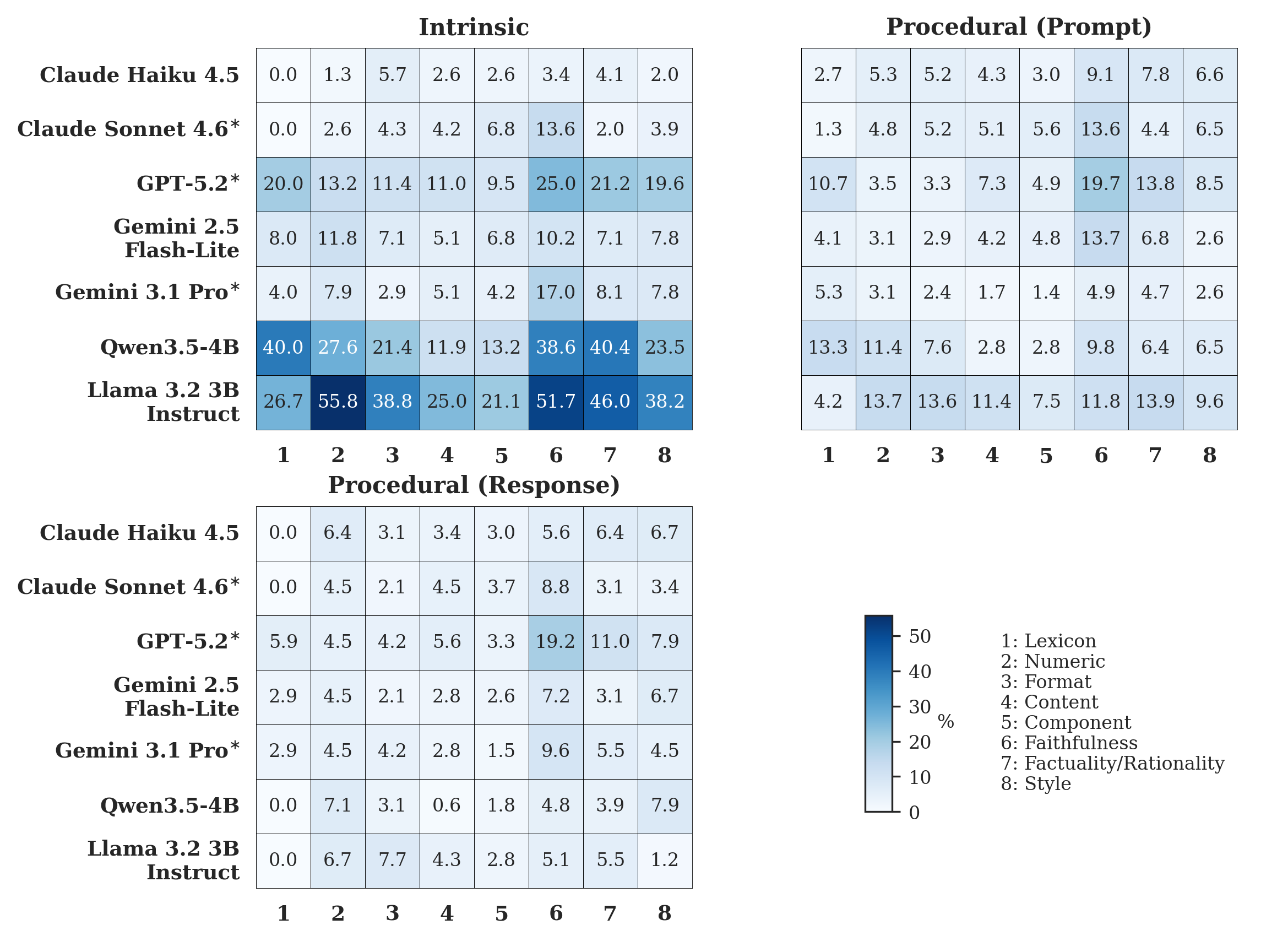}
    \caption{Constraint-level inconsistency rates by constraint type. Models marked with $^{\ast}$ are evaluated with reasoning enabled. Panels correspond to intrinsic inconsistency and procedural inconsistency under prompt-side and response-side perturbations.}
    \label{fig:inconsistency_by_constraint_type}
\end{figure*}

Figure~\ref{fig:inconsistency_by_constraint_type} shows that inconsistency varies across constraint categories and does not reduce to a single dominant failure mode. The clearest regularity appears in procedural inconsistency, where \textit{Faithfulness} is repeatedly among the most inconsistent categories for several proprietary judges, with \textit{Factuality/Rationality} elevated for some models. Intrinsic inconsistency is more model-dependent, with some models showing localized instability on particular categories, such as GPT-5.2 on \textit{Faithfulness} and \textit{Factuality/Rationality}, while the two open-source baselines show high inconsistency across a wider range of categories. Overall, the figure suggests that procedural inconsistency is more likely to arise on categories that require grounding or plausibility judgments, meaning that these more interpretive verification tasks appear less stable under prompt or response reformulation than surface-oriented checks.
\end{document}